\documentclass{article}
\usepackage{graphicx} % Required for inserting images
\usepackage{wrapfig}
\usepackage{subcaption}
\usepackage{caption}
\usepackage{hyperref}
\newcommand{\minisection}[1]{\vspace{0.005in} \noindent {\bf #1}}
\title{On the Occurence of Critical Learning Periods in Neural Networks
}
\author{Stanisław Pawlak}
\date{}

\begin{document}

\maketitle

\begin{abstract}
  This study delves into the plasticity of neural networks, offering empirical support for the notion that critical learning periods and warm-starting performance loss can be avoided through simple adjustments to learning hyperparameters. The critical learning phenomenon emerges when training is initiated with deficit data. Subsequently, after numerous deficit epochs, the network's plasticity wanes, impeding its capacity to achieve parity in accuracy with models trained from scratch, even when extensive clean data training follows deficit epochs. Building upon seminal research introducing critical learning periods, we replicate key findings and broaden the experimental scope of the main experiment from the original work. In addition, we consider a warm-starting approach and show that it can be seen as a form of deficit pretraining. In particular, we demonstrate that these problems can be averted by employing a cyclic learning rate schedule. Our findings not only impact neural network training practices but also establish a vital link between critical learning periods and ongoing research on warm-starting neural network training.
\end{abstract}

\section{Introduction}
Recent research on neural network training \cite{achille2019criticallearningperiodsdeep} indicates the existence of a crucial phase at the beginning of training, known as the 'critical learning period' during which the network learns patterns that limit its further adaptability, because they are challenging to overwrite. If true, this poses a significant challenge for continual learning (CL)~\cite{1999french, kirkpatrick2017overcoming, vandeven2019scenarios, hadsell2020embracing}, where learning objectives evolve over time, which can make CL training less secure~\cite{han2023data}. 
\begin{figure}[ht]
    \centering
     \begin{subfigure}{0.48\textwidth}
        \centering
         \includegraphics[width=\textwidth]{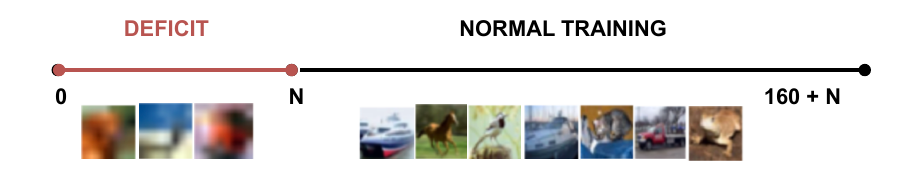}
    \end{subfigure}
         \begin{subfigure}{0.48\textwidth}
        \centering
        
    \includegraphics[width=\textwidth]{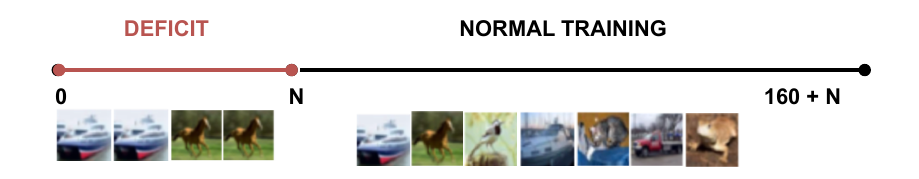}
    \end{subfigure}
     \begin{subfigure}{0.48\textwidth}
        \centering
         \includegraphics[width=\textwidth]{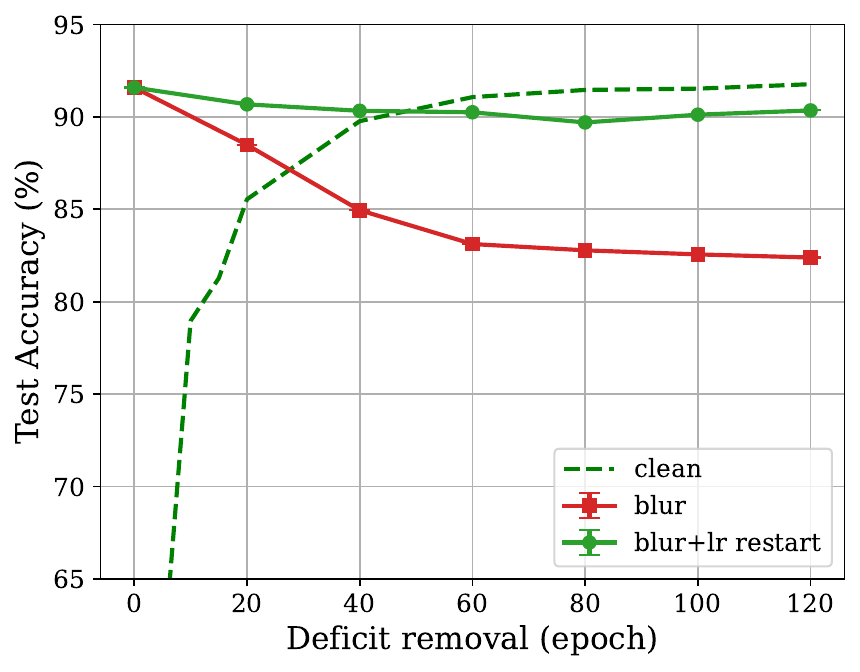}
          \caption{Critical learning periods}
    \end{subfigure}
         \begin{subfigure}{0.48\textwidth}
        \centering
        
    \includegraphics[width=\textwidth]{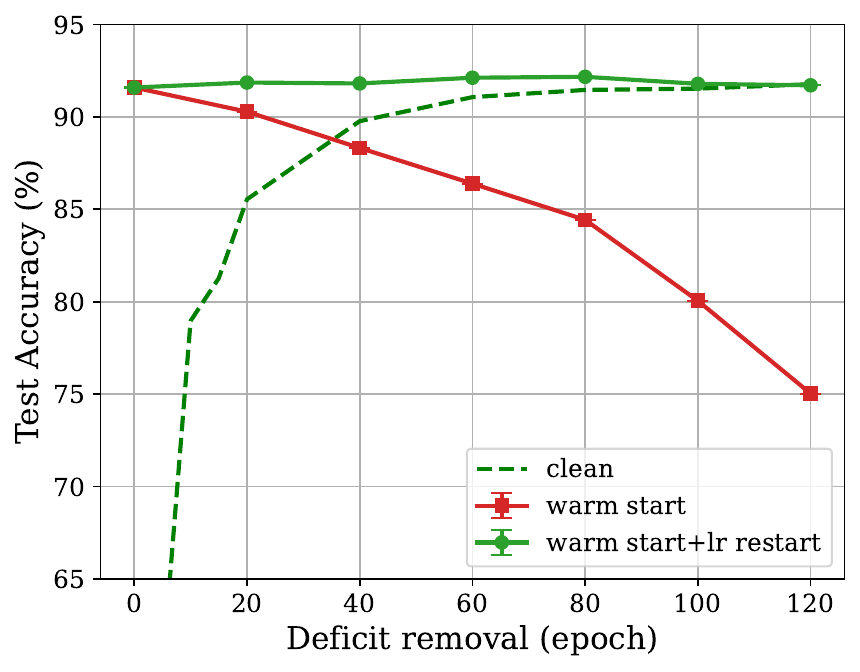}
    \caption{Warm-starting}
    \end{subfigure}
    \caption{\textbf{Restarting learning rate alleviate plasticity problems after deficit training epochs}. a) Restarting learning rate after deficit epochs almost closes the performance gap between deficit trained model and the model trained on clean data from scratch. b) It closes the performance gap when we use pretraining on a small subset of data (warm-starting) as a deficit during deficit epochs training. Image above the plot reproduced based on the original work~\cite{achille2019criticallearningperiodsdeep}.}
    \label{fig:teaser}
\end{figure}
\begin{figure}[ht]
    \centering
     \begin{subfigure}{0.48\textwidth}
        \centering
           \includegraphics[width=\textwidth]{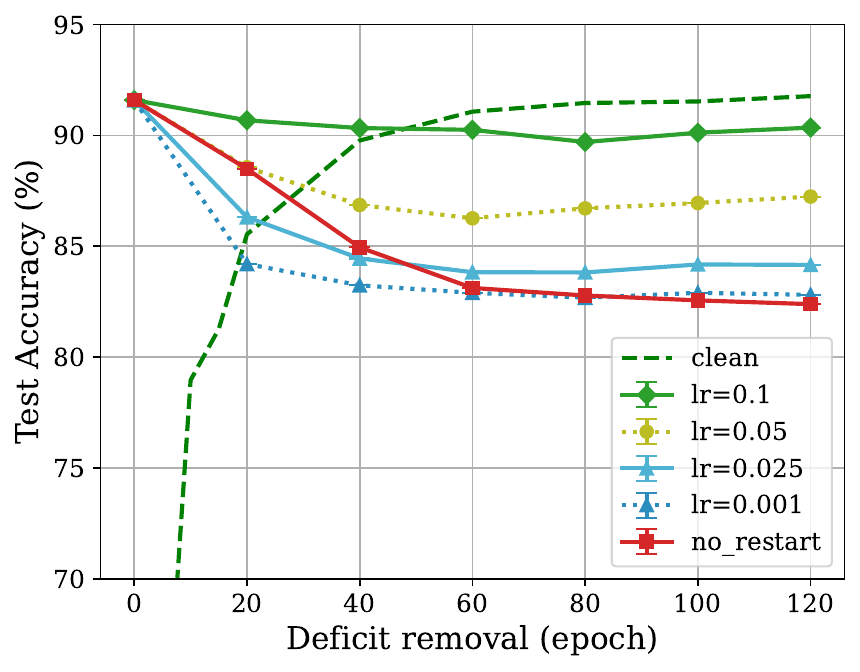}

          \caption{Critical learning periods}
    \end{subfigure}
         \begin{subfigure}{0.48\textwidth}
        \centering
        
    \includegraphics[width=\textwidth]{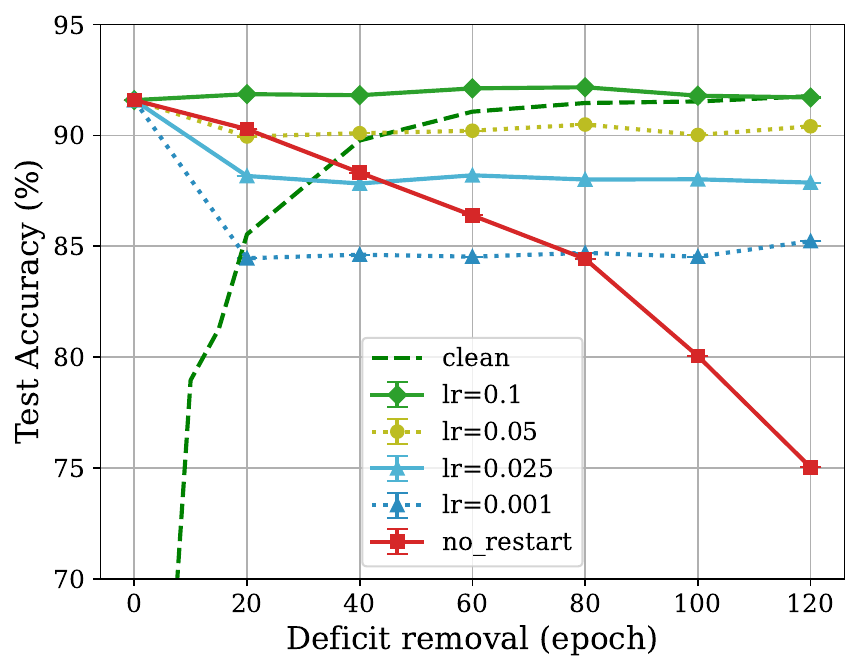}    \caption{Warm-starting}
    \end{subfigure}
    \caption{\textbf{Final model accuracy correlates with learning rate size after deficit epochs (restart lr).} In both cases the bigger the restarting lr, the better the final result. Also, the results are worse in situations where original learning rate is bigger than restarted after the last deficit epoch (e.g. 20 and 40 epoch for lr=0.001).}
    \label{fig:lr_restart}
\end{figure}
In fact, there is another avenue of works related to CL through the problem of plasticity loss when warm-starting neural network training~\cite{ash2020warmstartingneuralnetworktraining}. 
%Plasticity refers to the ability to learn and adapt to new data. 
In~\cite{ash2020warmstartingneuralnetworktraining} authors show that the plasticity of the network is affected by warm-starting: when the training starts with only a subset of total data, and the data is gradually extended during the training, the final performance of the network is lower compared to the standard training. 

In this work, we analyze both problems using combined experimental setup. Firstly, we reproduce and analyze the main experiment from~\cite{achille2019criticallearningperiodsdeep}, broaden the scope of the evaluated deficits, and investigate the impact of the severity of the deficit on performance loss. Secondly, we recreate the warm-starting experiment using exact~\cite{achille2019criticallearningperiodsdeep} setup and show that starting a training from a subset containing clean data can be seen as a kind of deficit training. Then, following ideas from~\cite{berariu2023studyplasticityneuralnetworks} we show that the plasticity loss problems presented previously can be almost fully avoided through simple adjustments to learning hyperparameters like incorporating cyclic learning rate. Finally, we consider and investigate targeted deficit training that influences only a subset of training data. Being more challenging to discover, it can lead to spurious correlations and biases in the decisions of the final trained model.

\section{Experiments}
\subsection{Experimental Setup}
Our experimental setup closely follows that of Achille et al.~\cite{achille2019criticallearningperiodsdeep}. In particular, we replicate the ResNet-based configuration of their main experiment, illustrated in Figure~\ref{fig:teaser}. We employ the ResNet-18 architecture~\cite{he2015deepresiduallearningimage}, trained using stochastic gradient descent (SGD) with a batch size of 128. The initial learning rate is set to 0.1 and decays exponentially by a factor of 0.97 per epoch, with a weight decay of 0.0005. For experiments using a fixed learning rate, we set it to 0.001.
All experiments are conducted on the CIFAR-10 dataset~\cite{alex2009learning}, augmented with standardized image corruptions from CIFAR-C~\cite{hendrycks2019robustness}. The blur deficit introduced in~\cite{achille2019criticallearningperiodsdeep} was found to substantially degrade image quality. Based on the description provided in that work, we reproduced this effect by extrapolating the CIFAR-C pixelate corruption to a severity level of 9 (note that CIFAR-C originally defines severity levels from 1 to 5).
\subsection{Results}
\subsubsection{Replication of phenomenons from critical learning periods and warm-starting.}
\minisection{Critical learning periods.} Figure~\ref{fig:teaser}a replicates the main experiment from Achille et al.~\cite{achille2019criticallearningperiodsdeep}. The top-left point denotes the accuracy of a ResNet-18 model trained on clean data for 160 epochs. The x-axis indicates the deficit removal epoch, i.e., the number of deficit pretraining epochs preceding 160 clean epochs on the full dataset. The green dashed line marks the accuracy of a clean model trained for the same number of epochs. Our results are about one percentage point lower than in the original work but reproduce the same qualitative behavior and performance drop, confirming the critical learning period effect.

\minisection{Warm-starting.} Figure~\ref{fig:teaser}b reproduces the warm-starting experiments using the same setup. Pretraining on 1,000 samples for $x$ deficit epochs consistently degrades final performance, even after 160 additional training epochs on the full dataset. Compared to Figure~\ref{fig:teaser}a, longer deficit pretraining causes a noticeably larger accuracy drop.

\begin{figure}[ht]
    \centering
     \begin{subfigure}{0.32\textwidth}
        \centering
           \includegraphics[width=\textwidth]{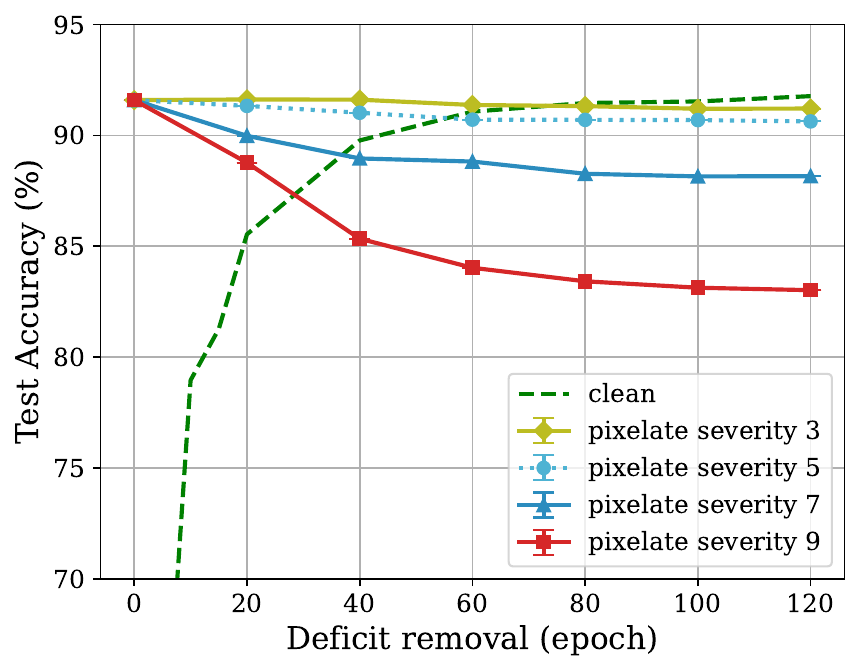}

          \caption{Pixelate}
    \end{subfigure}
     \begin{subfigure}{0.32\textwidth}
        \centering
           \includegraphics[width=\textwidth]{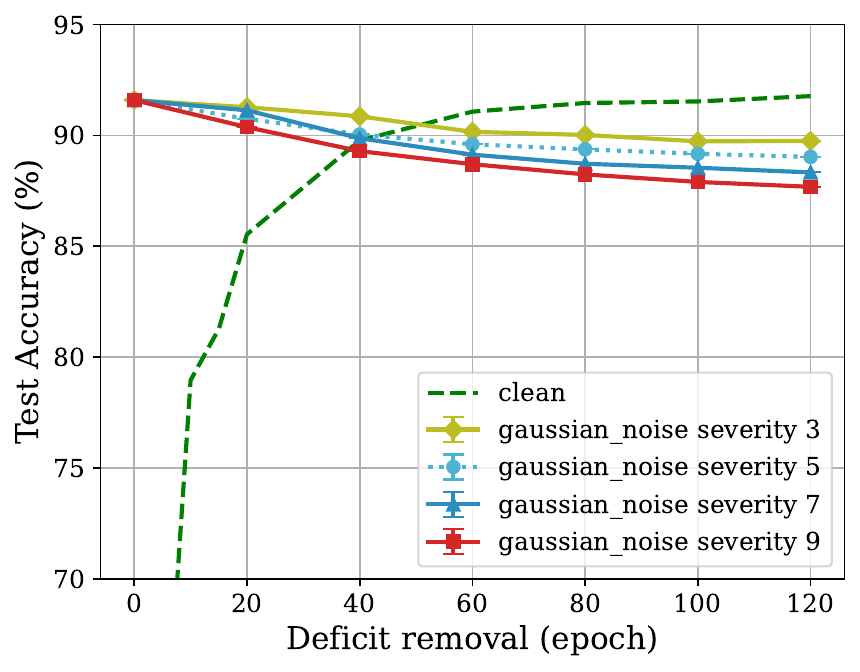}

          \caption{Gaussian noise}
    \end{subfigure}
     \begin{subfigure}{0.32\textwidth}
        \centering
           \includegraphics[width=\textwidth]{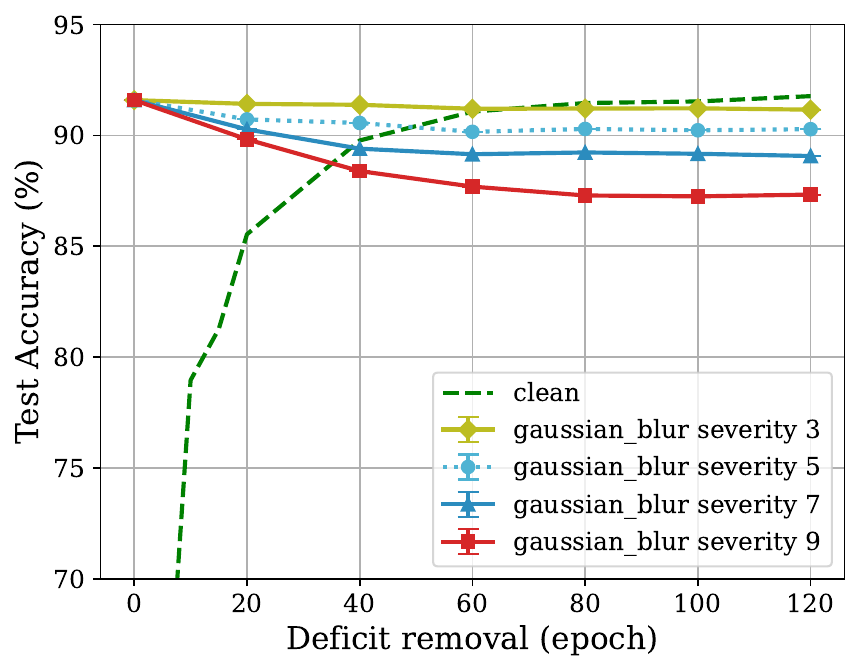}

          \caption{Gaussian blur}
    \end{subfigure}
    \caption{\textbf{Corruption severities correlates with performance gap size.} Bigger gap and higher sensitivity to corruption level is observed for some standardized corruptions types (pixelate $>$ gaussian blur $>$ gaussian noise).}
    \label{fig:severities}
\end{figure}

\subsubsection{Cyclic learning rate alleviates performance loss.}
Figure~\ref{fig:teaser} not only reproduces prior findings but also reveals a practical intervention: adjusting the learning rate after the deficit phase can substantially mitigate the resulting performance loss. When the learning rate is reset to its initial value, the performance gap observed in the warm-starting setup disappears entirely, and the degradation associated with critical learning periods is significantly reduced.

Figure~\ref{fig:lr_restart} provides a deeper look into this effect. We observe a clear correlation between final model accuracy and the restarted learning rate magnitude. Higher restart values—those exceeding the exponentially decayed rate from the 0.97 per-epoch schedule—consistently lead to improved recovery. In contrast, restarting from a smaller value further widens the gap, particularly when the deficit is removed at epochs 20 or 40 with a learning rate of 0.001.

These findings suggest that the learning rate plays a crucial role in enabling the network to recover from suboptimal early training. A low or fixed learning rate may trap the model in a narrow region of the loss landscape, preventing effective adaptation once clean data become available. Periodically increasing or resetting the learning rate appears to restore the model’s capacity for plasticity—allowing it to “relearn” after exposure to degraded data.

\subsubsection{Corruption severity correlates with performance loss.}
The previous results show that the effects of critical learning periods are not as catastrophic or irreversible as initially suggested. Nonetheless, this experimental setup remains valuable for probing how neural networks acquire and retain visual patterns during training. To explore this further, we extended the main experiment from~\cite{achille2019criticallearningperiodsdeep} to examine how varying corruption severities influence the performance gap, using the standard setup without cyclic learning rate adjustments.

We found that higher severity levels across three standardized image corruptions—pixelate, Gaussian noise, and Gaussian blur—consistently lead to larger performance drops (see Figure~\ref{fig:severities}). Interestingly, the extent of degradation depends on the spectral characteristics of the corruption. Corruptions emphasizing high-frequency noise (e.g., Gaussian noise) yield smaller performance gaps than those that suppress mid- and high-frequency components (e.g., pixelate and Gaussian blur). This pattern aligns with the known spectral bias of neural networks~\cite{spectral_bias}.

As shown by Achille et al.~\cite{achille2019criticallearningperiodsdeep} (Figure 8), low-frequency features acquired in early convolutional layers during deficit epochs are difficult to override during subsequent clean training. This rigidity increases with the severity of the corruption. For moderate corruptions (severity 5), pretraining with Gaussian noise results in the largest performance drop after 140 deficit epochs. However, at higher severities (e.g., 9), Gaussian noise becomes the least harmful among the examined corruptions. This trend indicates that restricting learning primarily to high-frequency information, as in Gaussian noise, is less damaging than distortions that eliminate both medium- and high-frequency components, such as Gaussian blur or pixelate. The results further reinforce the view that access to a balanced spectral range of visual information is crucial for robust feature acquisition during early training.

\begin{wrapfigure}{r}{0.32\textwidth}
\vspace{-0.0cm}
    \centering
    \includegraphics[width=0.32\textwidth]{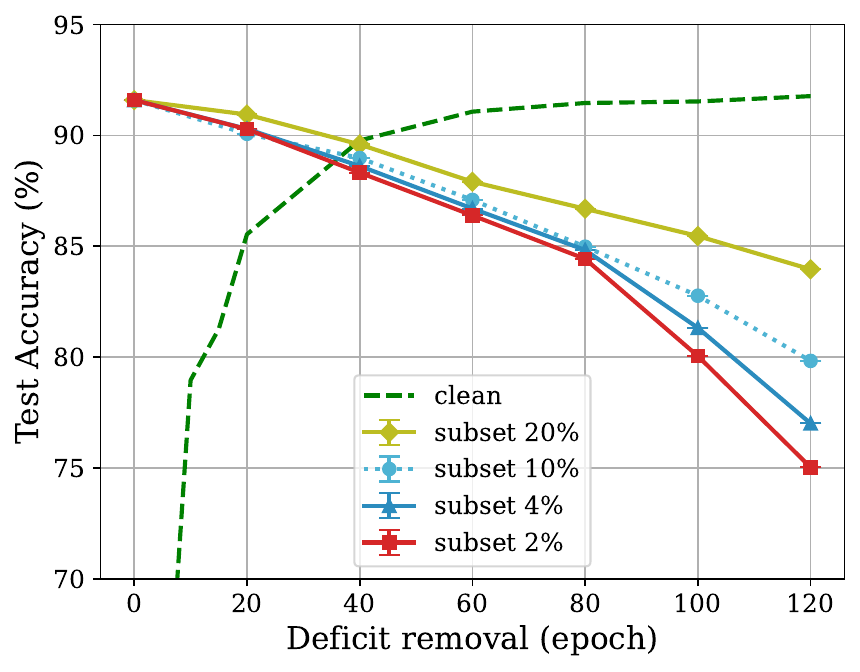}
    \caption{\textbf{Smaller subset size in deficit training increases the performance gap.}
    }
    \vspace{-0.3cm}
    \label{fig:subsets}
\end{wrapfigure}

We want to stress out that these results are valid as long as the corruptions are added to all images and are not overriding the whole information from the images. In the latter case, the original work shows that switching images with fixed noise without any correlation to labels deteriorates the performance less compared to limiting real information from images by corrupting them. In other words, if we keep increasing the severity of corruptions, we would finally obtain corrupted images with no useful information about classes, and thus the performance gap would be lower - similar to the one for the fixed noise images.
\subsubsection{Warm-starting as deficit training.}
Figure~\ref{fig:subsets} demonstrates that warm-starting with progressively smaller data subsets can be viewed as a form of deficit training with increasing corruption severity. When the subset is reduced to only 2\% of the training data, the resulting performance gap becomes substantially larger than that observed for the image corruptions discussed in the previous section. This suggests that overfitting to a limited subset of data impairs network plasticity more severely than exposure to heavily corrupted inputs.

\subsubsection{Expanding deficits with standard image corruptions.}
\begin{wrapfigure}{r}{0.5\textwidth} % 'r' for right placement, width set to 50% of the text width
    \centering
    \vspace{-0.4cm}
    \begin{minipage}{0.49\linewidth}
        \includegraphics[width=\linewidth]{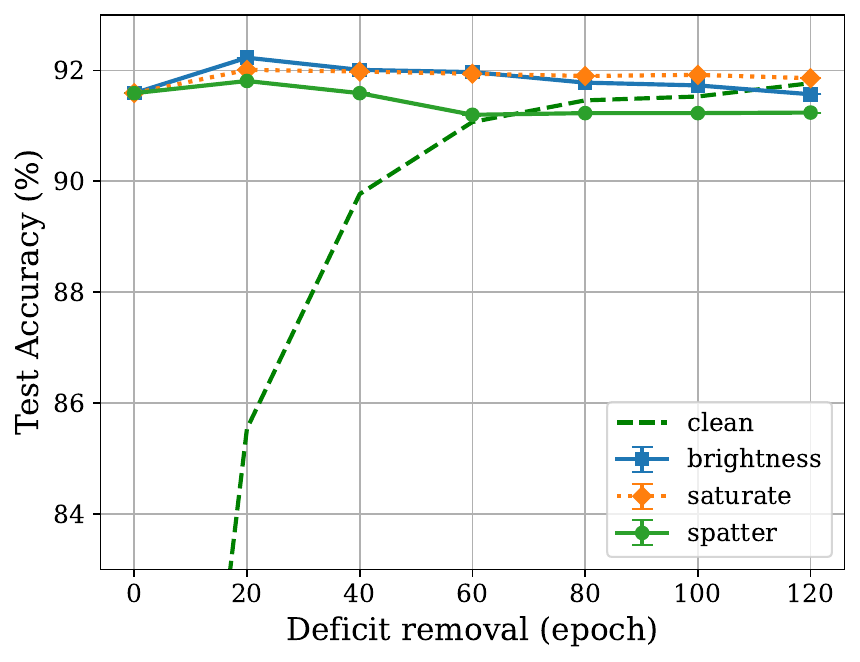}
        \caption*{Decreasing lr}
    \end{minipage}
    \hfill % Adds some space between the images
    \begin{minipage}{0.49\linewidth}
        \includegraphics[width=\linewidth]{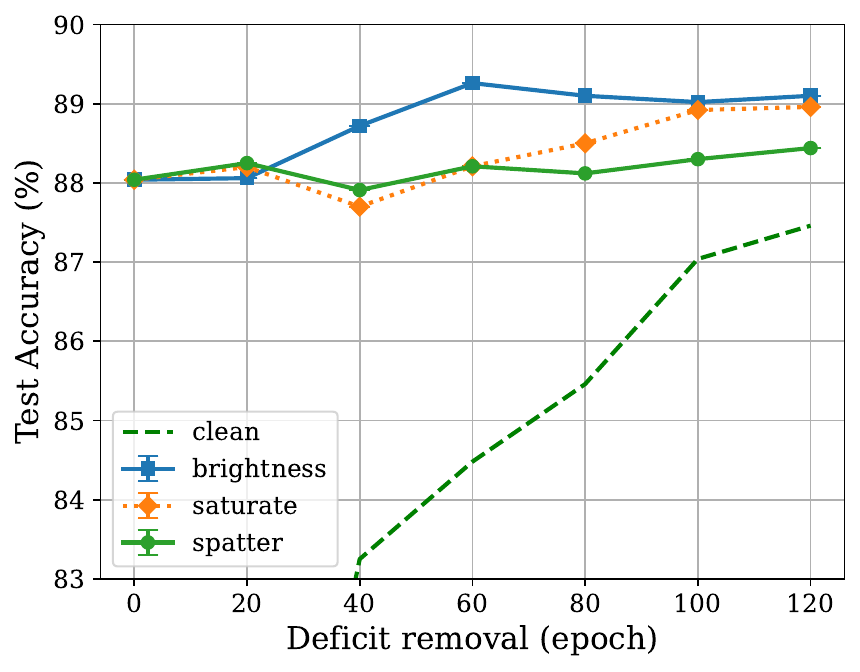}
        \caption*{Const lr}
    \end{minipage}
        \vspace{-0.2cm}
    \caption{\textbf{Mild corruptions: \\ no performance loss.}}
    \label{fig:mild_corr}
        \centering
    \begin{minipage}{0.49\linewidth}
        \includegraphics[width=\linewidth]{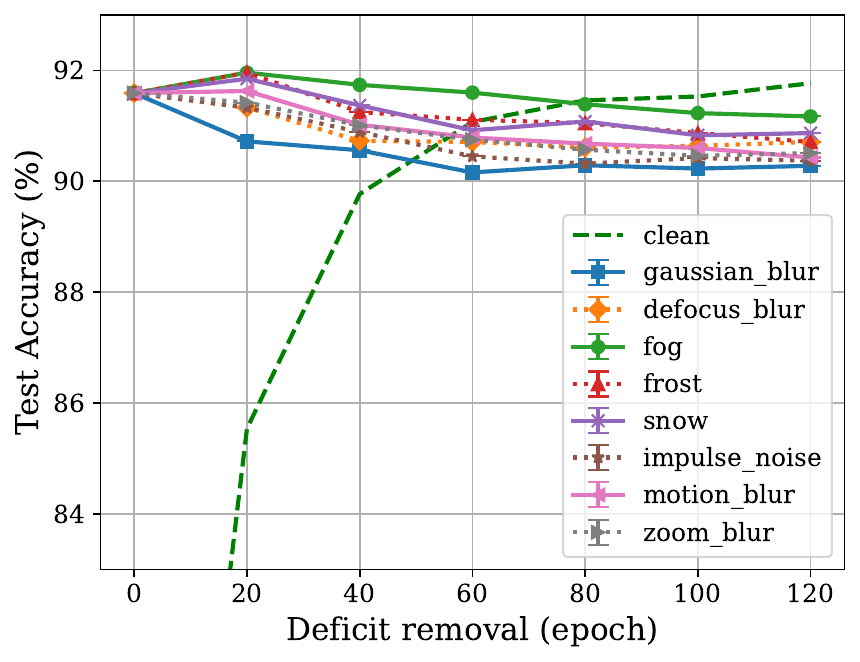}
       \caption*{Decreasing lr}
    \end{minipage}
    \hfill % Adds some space between the images
    \begin{minipage}{0.49\linewidth}
        \includegraphics[width=\linewidth]{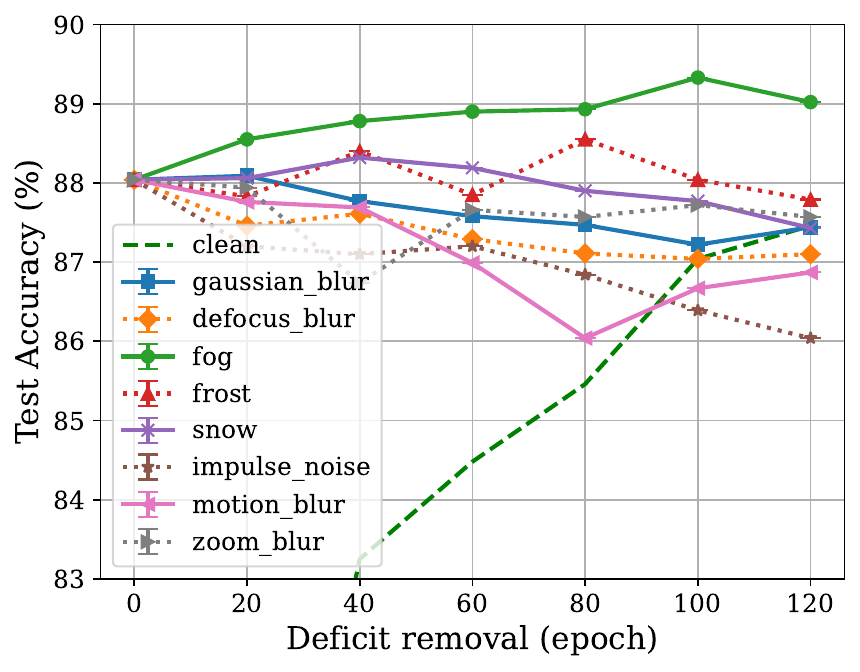}
        \caption*{Const lr}
    \end{minipage}
       \vspace{-0.2cm}
    \caption{\textbf{Medium corruptions: small performance loss.}}
     \label{fig:medium_corr}
         \centering
    \begin{minipage}{0.49\linewidth}
        \includegraphics[width=\linewidth]{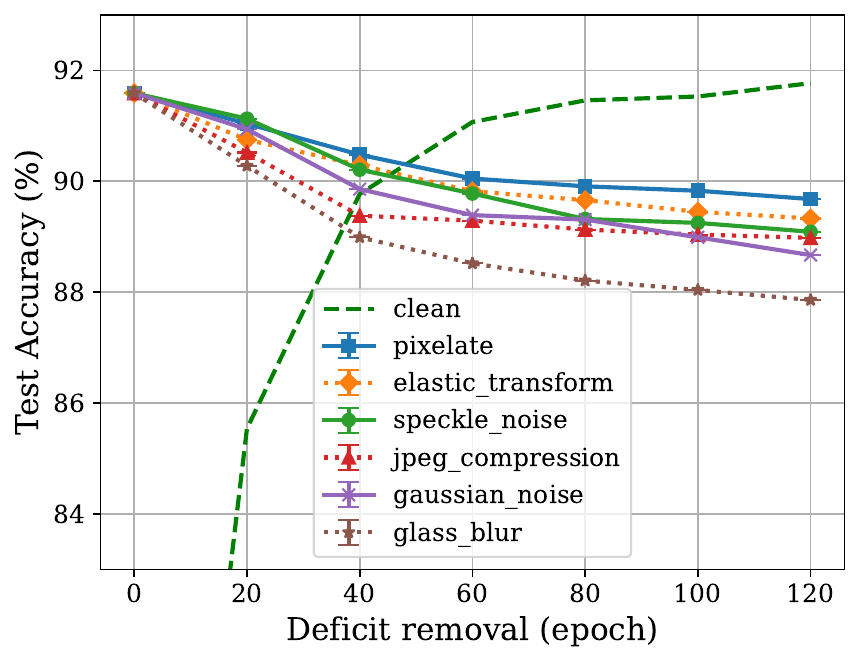}
       \caption*{Decreasing lr}
    \end{minipage}
    \hfill % Adds some space between the images
    \begin{minipage}{0.49\linewidth}
        \includegraphics[width=\linewidth]{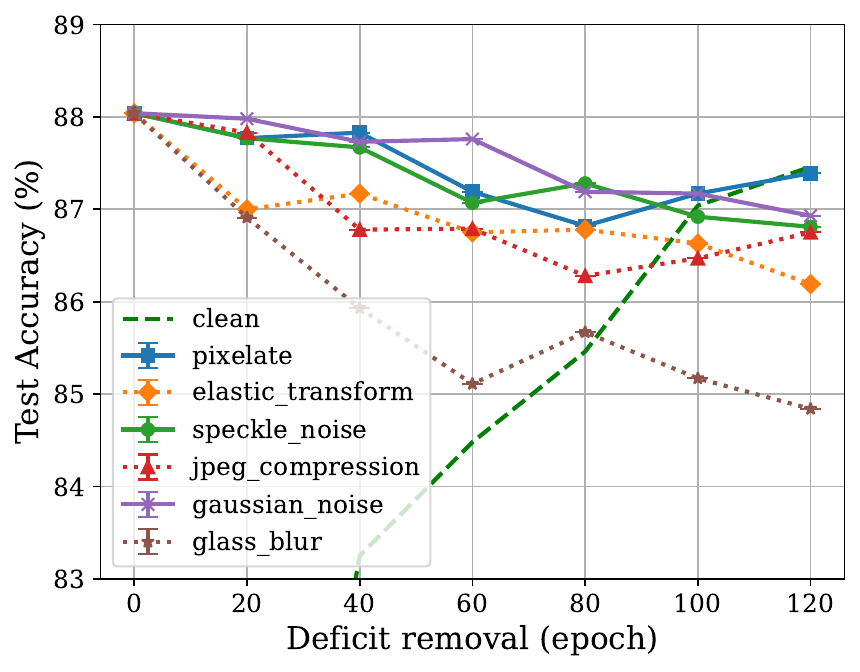}
        \caption*{Const lr}
   
    \end{minipage}
        \vspace{-0.2cm}
    \caption{\textbf{Strong corruptions: \\ significant performance loss.}}
    \label{fig:strong_corr}
       % \vspace{-0.6cm}
\end{wrapfigure}

Figures~\ref{fig:mild_corr}, \ref{fig:medium_corr}, and~\ref{fig:strong_corr} illustrate the impact of different corruption types when the severity level is fixed at a moderate value of five. We categorize the corruptions into three groups based on the performance gap observed after deficit training:

\begin{itemize}
\item \textit{Mild-type corruptions} (Figure~\ref{fig:mild_corr}) — no measurable performance gap is observed: \textit{brightness, saturate, spatter}.
\item \textit{Medium-type corruptions} (Figure~\ref{fig:medium_corr}) — a performance gap is present but remains below 2 percentage points under the decreasing learning rate schedule: \textit{gaussian blur, defocus blur, fog, frost, snow, impulse noise, motion blur, zoom blur}.
\item \textit{Strong-type corruptions} (Figure~\ref{fig:strong_corr}) — a clear performance gap exceeding 2 percentage points is observed under the decreasing learning rate schedule: \textit{pixelate, elastic transform, speckle noise, jpeg compression, gaussian noise, glass blur}.
\end{itemize}

\textit{Mild-type} corruptions, such as brightness adjustments, act as light data augmentation during pretraining. With a fixed learning rate, they can even yield slightly better results than clean training.
\textit{Medium-type} corruptions mainly preserve low-frequency content, capturing general features like shape or background color. At severity level five, their impact is moderate, though the performance gap increases markedly with higher severity levels.
\textit{Strong-type} corruptions produce large performance drops even when the overall image structure remains intact. These typically interfere with high-frequency patterns (e.g., Gaussian noise, speckle noise) or disrupt local image microstructure (e.g., glass blur).

Overall, performance degradation depends not only on corruption severity but also on which visual information is distorted. Corruptions that disrupt spatial structure or frequency balance (e.g., pixelate, glass blur) hinder early feature learning more severely than those altering only color or intensity.

\subsubsection{Targeted deficit training.}
Figures~\ref{fig:teaser} and~\ref{fig:lr_restart} show that applying a cyclic learning rate can almost completely eliminate the performance gap observed in critical learning periods. However, increasing the learning rate also amplifies the magnitude of weight updates, leading to more substantial model changes. While this is effective for full retraining, it may not be ideal when the deficit affects only a subset of classes. In such cases, inconsistencies in model predictions are harder to detect, as errors remain localized to those specific categories.

\begin{figure}[bt]
    \centering
    \includegraphics[width=\textwidth]{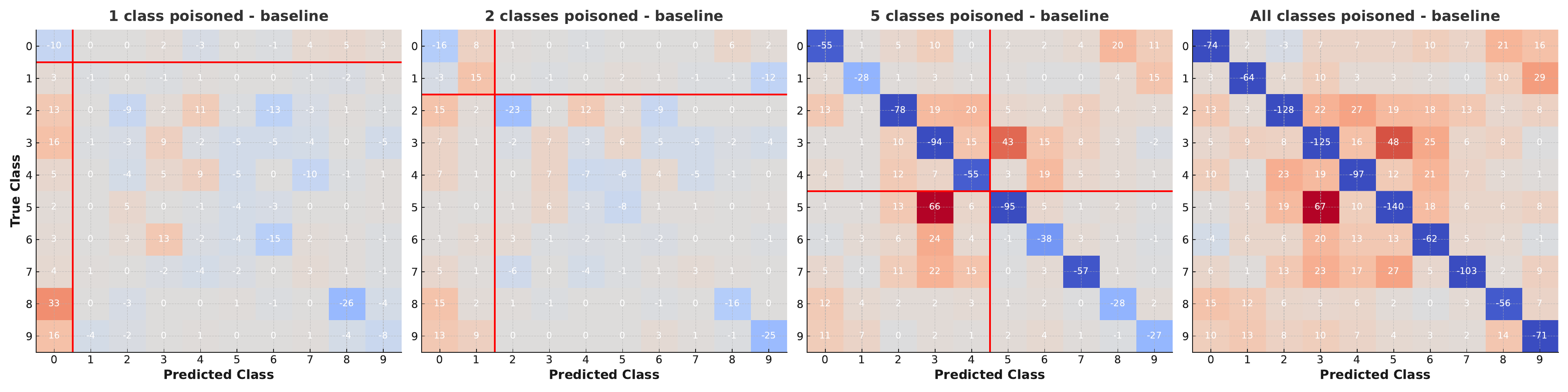}
    \caption{\textbf{Differences in confusion matrices for runs with deficit data only for 1,2,5,10 classes.} Deficit classes are on the left and up to the red lines. Clean class examples are frequently misclassified as belonging to deficit-pretrained classes.}
    \label{fig:conf_matrices}
\end{figure}

Figure~\ref{fig:conf_matrices} visualizes this effect using confusion matrices that show the difference between models trained with and without deficits. In these experiments, only a subset of images belonging to particular classes was corrupted during the deficit phase. The affected classes, highlighted in the top-left corner by red lines, exhibit pronounced confusion with visually similar categories. This indicates that targeted deficits can lead to class-dependent forgetting and representational imbalance, even when the overall accuracy remains high.

We further observe that the number of misclassifications increases with the number of classes exposed to deficit data. When only one class is affected, it alone tends to produce numerous false predictions. As the number of deficit classes grows -- e.g., when half of the classes are corrupted -- the classifier struggles to distinguish between deficit and clean categories. For example, samples from classes 0–4 are frequently misclassified as 5–9 and vice versa. In contrast, no such confusions occur among the always-clean classes (the lower-right corner of the matrix).

\section{Conclusion}
In this work, we provide empirical evidence that performance degradation observed during critical learning periods and warm-starting can be mitigated through simple adjustments to learning hyperparameters. We extend the original experimental framework by exploring the effects of cyclic learning rates, as well as various types and severities of image corruptions. Additionally, we reinterpret warm-starting as a form of deficit pretraining and show that targeted deficits can affect only a subset of classes -- an effect that is harder to detect and less responsive to cyclic learning rate adjustments.
\newpage
Overall, our findings bridge two previously distinct research directions on neural network plasticity, linking critical learning periods with warm-starting strategies. This connection underscores the importance of training dynamics and learning rate scheduling in preserving model adaptability during learning.
\bibliographystyle{abbrv}
\bibliography{bibliography.bib}

\end{document}